\PassOptionsToPackage{table,xcdraw}{xcolor}
\documentclass{arxiv_preprint} 

\usepackage{amsmath,amsfonts,bm}

\def\eqref#1{equation~\ref{#1}}

\def\1{\bm{1}}

\DeclareMathAlphabet{\mathsfit}{\encodingdefault}{\sfdefault}{m}{sl}
\SetMathAlphabet{\mathsfit}{bold}{\encodingdefault}{\sfdefault}{bx}{n}

\usepackage{algorithm}
\usepackage{algpseudocode}
\usepackage{listings}

\hypersetup{
    pdftitle={On-Policy Parameter Update Direction Underlies Generalization in LLM Post-Training},
    pdfauthor={TODO: authors},
}

\lstdefinestyle{pseudocode}{
    language=Python,
    basicstyle=\ttfamily\small,
    keywordstyle=\color{blue!70!black}\bfseries,
    commentstyle=\color{green!45!black}\itshape,
    stringstyle=\color{orange!70!black},
    numbers=left,
    numberstyle=\tiny\color{gray},
    numbersep=8pt,
    frame=single,
    rulecolor=\color{black!30},
    backgroundcolor=\color{gray!4},
    breaklines=true,
    breakatwhitespace=true,
    showstringspaces=false,
    columns=fullflexible,
    keepspaces=true,
    tabsize=4,
    captionpos=b
}

\newcommand{\corrauth}{\text{\raisebox{-0.12ex}{\scalebox{0.78}{\faEnvelope}}}}

\title{On-Policy Parameter Update Direction Underlies Generalization in LLM Post-Training}

\author[1,2]{Shufan Shen}
\author[3,\corrauth]{~Zhongni Hou}
\author[1,2]{~Junshu Sun}
\author[3]{~Yufei Zhang}
\nextauthorline
\author[3]{Wei Lin} 
\author[3]{~Guojun Yin}
\author[1,2]{~Qingming Huang}
\author[1,\corrauth]{~Shuhui Wang}

\affiliation[1]{State Key Lab. of AI Safety, Institute of Computing Technology, Chinese Academy of Sciences}
\nextaffiliationline
\affiliation[2]{University of Chinese Academy of Sciences}
\affiliation[3]{Meituan}

\contribution[\corrauth]{Corresponding authors}

\metadata[\raisebox{-0.18ex}{\scalebox{0.89}{\faEnvelope}}~Contact]{\email{shenshufan22z@ict.ac.cn}, \email{wangshuhui@ict.ac.cn}}
\metadata[\raisebox{-0.14ex}{\scalebox{0.92}{\faGithub}}~Code]{\url{https://github.com/ssfgunner/OPSFT}}

\begin{document}

\begin{abstract}
The strong generalization performance of on-policy post-training paradigms has motivated studies of their parameter update behaviors. 
However, these studies treat the observed behaviors only as byproducts in on-policy training, overlooking their potential to serve as optimization principles for improving the generalization of other paradigms such as supervised fine-tuning~(SFT).
To address this limitation, we investigate whether there exists a specific on-policy update behavior that can achieve such improvements.
First, our theoretical and experimental analyses reveal that SFT updates parameters along consistent directions, while the on-policy paradigm continuously adjusts the direction during training. 
This difference inspires us to focus on the cumulative update direction of each parameter as a promising behavior. Then, we evaluate its effectiveness for improving generalization by proposing On-Policy direction-constrained Supervised Fine-Tuning~(OPSFT), which constrains SFT updates to the direction identified by on-policy paradigms. The strong performance of OPSFT indicates that the generalization advantage of on-policy paradigms can be transferred to SFT through the parameter update direction. Once such a direction is identified, even SFT can generalize with its updates constrained to this direction. This finding offers two practical benefits by combining the strong generalization of on-policy paradigms with the advantages of SFT, including the high training efficiency and ability to leverage high-quality trajectories.
For efficiency, we identify update directions that support strong generalization using a few on-policy training steps, and subsequently apply OPSFT to achieve high training efficiency.
For leveraging high-quality trajectories, OPSFT can utilize these trajectories to continue improving a post-trained model along its update direction without disrupting the ability learned from on-policy training.
\end{abstract}

\maketitle

\section{Introduction}
The on-policy post-training has emerged as an important paradigm for enhancing the reasoning capabilities of large language models~(LLMs)~\citep{shao2024deepseekmath, lu2025onpolicydistillation, agarwal2024policy}. 
Compared to the off-policy paradigm such as supervised fine-tuning~(SFT) that optimizes the model using fixed ground-truth responses~\citep{hinton2015distilling,gu2024minillm}, the on-policy paradigm optimizes the model using its self-generated responses and achieves strong generalization performance across diverse tasks~\citep{zhang2025100, guo2025deepseek, xiaomi2025mimo}.

The strong generalization of on-policy paradigms has motivated studies of their internal mechanisms to understand the reasons behind their success~\citep{nguyen2025reasoning,mukherjee2026reinforcement}. 
Prior studies investigate the effects of individual components that distinguish on-policy paradigms from SFT, such as the reverse KL divergence in OPD~\citep{nguyen2025reasoning} and negative samples in GRPO~\citep{abdolmaleki2025learning}. Despite these efforts, on-policy paradigms and SFT differ in numerous components~\citep{zhao2026large}, making component-wise analysis costly and fragmented. 
To obtain more unified conclusions, recent studies bypass individual components and focus directly on their resulting parameter updates. 
By comparing the updates of on-policy paradigms and SFT from both the spectrum~\citep{wu2025invisible, zhu2025path, shen2026geometry} and weight matrix~\citep{mukherjee2026reinforcement, zhang2026geora} perspectives, these studies provide insights into the distinctive parameter update behaviors of on-policy paradigms, such as the sparse update locations~\citep{mukherjee2026reinforcement} and the small spectrum shift~\citep{shen2026geometry}.

However, current studies treat these optimization behaviors only as byproducts of on-policy training~\citep{shen2026geometry, wu2025invisible}, rather than exploring their contributions to generalization. This perspective overlooks the potential of these behaviors to serve as effective optimization principles for improving the generalization of SFT.
Therefore, there remains a substantial gap between understanding the parameter update behaviors of on-policy paradigms and translating these insights into practical improvements in generalization performance.

\begin{figure}[tb]
  \centering
  \includegraphics[width=1.0\linewidth]{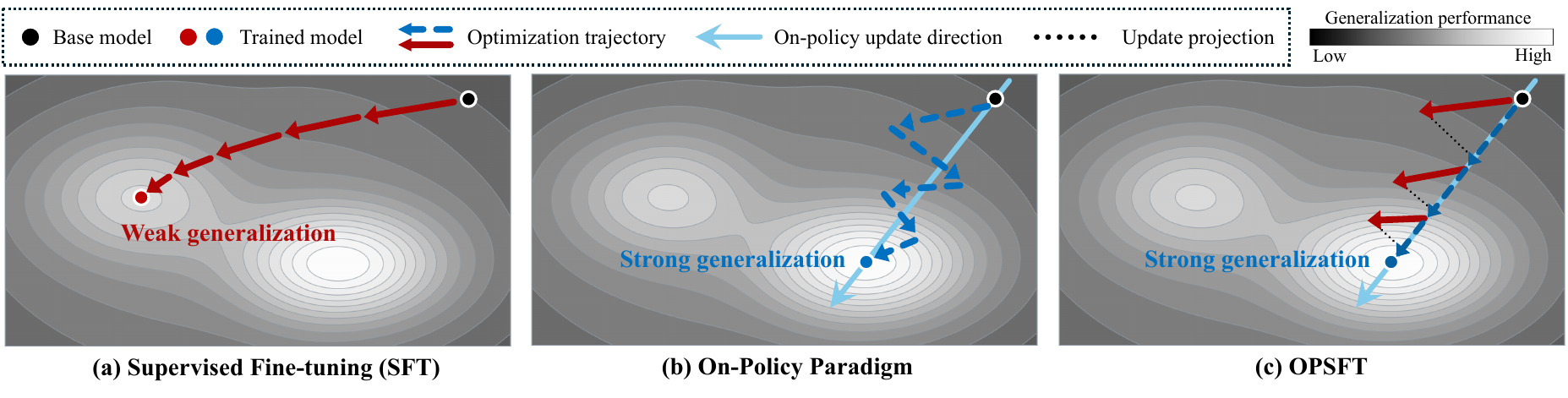}
  \vspace{-0.45cm}
  \caption{(a) SFT update parameters along consistent directions during training. (b) In contrast, on-policy paradigms continuously adjust the update direction. (c) To investigate whether the update direction identified by on-policy paradigms can improve the generalization of SFT, we introduce OPSFT that constrains the SFT update of each parameter to the on-policy direction. 
  }
  \label{fig:introduction}
  \vspace{-0.45cm}
\end{figure}

To bridge this gap, we investigate whether there exists an on-policy parameter update behavior that can improve the generalization of SFT.
First, inspired by studies of update locations~\citep{mukherjee2026reinforcement}, we extend the investigation target to update directions, which include both the location and sign of each parameter update. Our theoretical and experimental analyses demonstrate that, unlike SFT that updates along consistent directions~(Figure~\ref{fig:introduction}a), the on-policy paradigm continuously adjusts the update direction throughout training~(Figure~\ref{fig:introduction}b).
This continual effort for direction adjustment motivates us to investigate the resulting cumulative update direction of on-policy training as a promising behavior. 
To evaluate its effectiveness, we propose On-Policy direction-constrained SFT~(OPSFT), which retains SFT while constraining the update direction of each parameter to that identified by the on-policy paradigm~(Figure~\ref{fig:introduction}c). 
By comparing the generalization performance of OPSFT with vanilla SFT and on-policy paradigms, we evaluate the ability of the on-policy update direction to improve SFT. 
Experiments across different models~\citep{yang2025qwen3, guo2025deepseek} and benchmarks~\citep{Li2023HaluEvalAL, Zhou2023InstructionFollowingEF,aime24, aime25, cui2025process, he2026deepmath, balunovic2026matharena} demonstrate that OPSFT substantially outperforms vanilla SFT while achieving performance comparable to the on-policy paradigm.
The results suggest that the parameter update direction identified by on-policy paradigms can effectively transfer their generalization advantage to SFT. Once this direction is identified, even SFT can achieve strong generalization performance when its updates are restricted to this direction.

This finding benefits post-training by integrating the strong generalization of on-policy paradigms with the advantages of SFT, including high training efficiency and the ability to leverage high-quality trajectories.
For efficiency, we first conduct a few steps of on-policy training to identify the update direction and then apply OPSFT. By optimizing along the on-policy update direction without requiring continuous rollouts, we can achieve strong generalization and high training efficiency simultaneously.
For leveraging high-quality trajectories, given a model post-trained with an on-policy paradigm, OPSFT can continue training this model along its on-policy update direction, thereby avoiding disruption to the capabilities learned during on-policy training. This strategy allows the model to leverage newly acquired trajectories for continuous performance improvement without repeating the entire SFT-then-RL process.
Our contributions are illustrated as follows:
\begin{itemize}
    \item We analyze the parameter update direction of on-policy paradigms and reveal their preference for continuously adjusting the update direction during training.
    \item We propose OPSFT that constrains updates of SFT to the on-policy direction. Its strong performance indicates that the update direction identified by on-policy paradigms can transfer their generalization advantage to SFT.
    \item We leverage the investigation results to achieve strong generalization with high training efficiency, and utilize newly acquired high-quality trajectories to continue improving models post-trained by on-policy paradigms.
\end{itemize}

\section{Related Work}

\noindent\textbf{Optimization Behavior of On-Policy Post-Training}.
To understand the mechanisms behind the effectiveness of on-policy post-training, prior studies analyze the principal components of its optimization geometry~\citep{wu2025invisible, zhu2025path} and the impact of its training objectives on optimization behaviors~\citep{nguyen2025reasoning, abdolmaleki2025learning}. More fundamentally, recent studies directly investigate parameter updates and discover that on-policy updates are localized within small sub-networks~\citep{mukherjee2026reinforcement}.
Unlike existing studies that focus only on update locations and treat them as byproducts of on-policy paradigms~\citep{yu2026dense}, we extend the investigation to the update direction and discover its ability to transfer the generalization advantage of on-policy paradigms to SFT.

\noindent\textbf{Task-Relevant Parameter Localization}. Sparsely updating parameters is widely adopted to avoid overfitting in fine-tuning tasks or improve interpretability~\citep{zhang2024gradient, ansell2024scaling, gao2025weight, shen2025enhancing, shen2026vl}.
Current methods typically obtain the relevance between parameters and tasks and only adjust the task-relevant parameters throughout training~\citep{han2024sltrain, he2025smt}. Unlike prior approaches that rely on heuristic strategies under off-policy paradigms to estimate task relevance~\citep{he2023sensitivity, shen2024expanding, shen2026kernelized}, we demonstrate that the on-policy post-training paradigm naturally serves as an effective mechanism for identifying not only task-relevant parameter update locations but also directions. 

\noindent\textbf{Generalization Ability of Supervised Fine-tuning}. 
As a standard post-training approach, supervised fine-tuning~(SFT) is widely adopted for its simplicity and efficiency in imitating expert demonstrations~\citep{wu2026generalization, ren2026rethinking, xu2026adaptive}. However, its generalization performance falls significantly short of on-policy paradigms, such as GRPO~\citep{shao2024deepseekmath} and OPD~\citep{lu2025onpolicydistillation}. 
To bridge this gap, existing methods aim to enhance SFT by incorporating importance-sampling data selection strategies~\citep{qin2025supervised}, logit-weighted cross-entropy losses~\citep{wu2026generalization}, or negative samples~\citep{abdolmaleki2025learning}. Unlike these methods that imitate on-policy paradigms with training objectives and dataset constructions, we leverage the parameter update direction identified by on-policy paradigms to improve the generalization of SFT.

\section{On-Policy Update Direction Underlies Generalization}
In this section, we explore whether there exists an on-policy parameter update behavior that can transfer the generalization advantage of on-policy paradigms to SFT. Inspired by previous studies that focus on parameter update locations~\citep{shen2026geometry,mukherjee2026reinforcement}, we extend the investigation to the direction~(\textit{i.e.}, sign) and its evolution throughout training. 

\subsection{Analysis of Parameter Update Directions}

\noindent\textbf{Theoretical Analysis}. We compare gradient formulations of the on-policy paradigms and SFT to illustrate their different behaviors in parameter update direction at each training step.
Given an input prompt $x$, a response trajectory $\tau$, and the current policy $\pi_{\bm{\theta}}$ with parameters
$\bm{\theta}\in\mathbb{R}^{d}$, the gradient of the trajectory log-probability with respect to parameters $\bm{\theta}$ is computed as follows:
\begin{equation}
s_{\bm{\theta}}(x,\tau)
=
\nabla_{\bm{\theta}}\log\pi_{\bm{\theta}}(\tau\mid x).
\label{eq:trajectory_score}
\end{equation}
For every input prompt $x$, $s_{\bm{\theta}}(x,\tau)$ has zero conditional expectation:
\begin{equation}
\mathbb{E}_{\tau\sim\pi_{\bm{\theta}}(\cdot\mid x)}
\left[s_{\bm{\theta}}(x,\tau)\right]
=
\sum_{\tau}\pi_\theta(\tau\mid x)
\nabla_\theta\log\pi_\theta(\tau\mid x) 
=
\nabla_\theta\sum_{\tau}\pi_\theta(\tau\mid x)
=
\bm{0}_d,
\label{eq:score_zero_mean}
\end{equation}
where $\bm{0}_d\in\mathbb{R}^d$ denotes the zero vector.
According to the derivations of previous methods~\citep{shao2024deepseekmath}, the policy gradient of on-policy paradigms can be represented as follows,
\begin{equation}
g_{\mathrm{on}}(\bm{\theta})
=
\mathbb{E}_{x,\ \tau\sim\pi_{\theta}(\cdot\mid x)}
\left[
A_{\bm{\theta}}(x,\tau)
s_{\bm{\theta}}(x,\tau)
\right],
\label{eq:rl_gradient}
\end{equation}
where $A_{\bm{\theta}}(x,\tau)\in\mathbb{R}$ is the advantage of trajectory $\tau$ for prompt $x$ under the policy parameterized by $\bm{\theta}$. 
The projection of $g_{\mathrm{on}}(\bm{\theta})$ onto an arbitrary sign vector $\bm{v}\in\{-1,0,1\}^d$ can be formulated as:
\begin{equation}
\bm{v}^\top g_{\mathrm{on}}(\bm{\theta})=\mathbb{E}_{x,\ \tau\sim\pi_{\bm{\theta}}(\cdot\mid x)}
\left[
\bm{v}^\top A_{\bm{\theta}}(x,\tau)
s_{\bm{\theta}}(x,\tau)
\right].
\label{eq:direction}
\end{equation}
Given the zero conditional expectation of $s_{\bm{\theta}}(x,\tau)$ in Equation~\ref{eq:score_zero_mean}, Equation~\ref{eq:direction} can be reformulated as the projection onto \(\bm v\) of the covariance between \(A_{\bm{\theta}}(x,\tau)\) and \(s_{\bm{\theta}}(x,\tau)\):
\begin{equation}
\begin{aligned}
\bm{v}^\top g_{\mathrm{on}}(\bm{\theta})
=
\mathbb{E}_{x}
\big[
\bm{v}^\top \operatorname{Cov}_{\tau\sim\pi_{\bm{\theta}}(\cdot\mid x)}
\left[
A_{\bm{\theta}}(x,\tau), s_{\bm{\theta}}(x,\tau)
\right]
\big],
\end{aligned}
\label{eq:reward_score_covariance_derivation}
\end{equation}
For SFT, the projection of their gradients along $\bm{v}$ is:
\begin{equation}
\bm{v}^\top g_{\mathrm{sft}}(\bm{\theta})
=
-
\mathbb{E}_{x}
\big[
\bm{v}^\top \mathbb{E}_{\tau\sim \pi_{\mathrm{teacher}}(\cdot\mid x)}\left[s_{\bm{\theta}}(x,\tau)\right]
\big].
\label{eq:sft_gradient}
\end{equation}
By comparing Equation~\ref{eq:reward_score_covariance_derivation} with Equation~\ref{eq:sft_gradient}, we find that SFT updates parameters along the direction of $s_{\bm{\theta}}(x,\tau)$, where the trajectories $\tau$ are typically sampled from a fixed distribution $\pi_{\mathrm{teacher}}$. In contrast, on-policy paradigms update parameters toward the direction of the covariance matrix between the advantages $A_{\bm{\theta}}(x,\tau)$ and the trajectory gradients $s_{\bm{\theta}}(x,\tau)$. Since the trajectories $\tau$ are sampled from the current policy $\pi_{\bm{\theta}}$ that evolves during training, the distribution of advantages $A_{\bm{\theta}}(x,\tau)$ changes with $\pi_{\bm{\theta}}$ rather than serving as a fixed target. Consequently, the parameter update direction of on-policy paradigms changes with the evolving advantage distribution throughout training.

\begin{figure}[tb]
  \centering
  \includegraphics[width=1.0\linewidth]{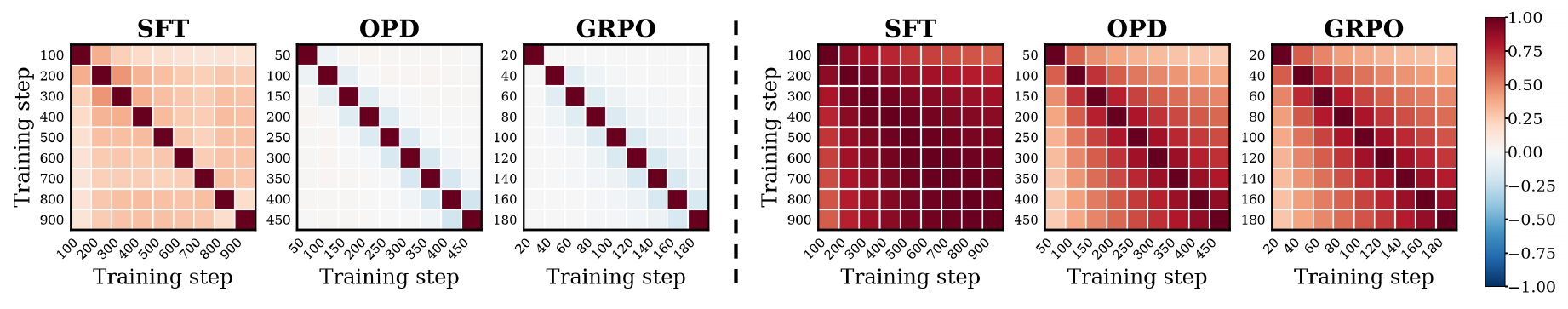}
  \vspace{-0.7cm}
  \caption{Comparisons of update directions between SFT and on-policy~(OPD, GRPO) paradigms using Qwen3-1.7B on DeepMath. We present the cosine similarity of interval~\textit{(left)} and cumulative~\textit{(right)} parameter updates among different training stages. 
  }
  \label{fig:prestudy_direction}
  \vspace{-0.3cm}
\end{figure}

\noindent\textbf{Experimental Analysis}. To verify the above theoretical analysis, we measure the cosine similarity among update directions at different training steps from both interval and cumulative perspectives, as shown in Figure~\ref{fig:prestudy_direction}. For interval updates, SFT optimizes towards positively correlated directions, while on-policy paradigms explore nearly orthogonal directions at different training stages. This observation is consistent with our analysis in Equation~\ref{eq:reward_score_covariance_derivation} and Equation~\ref{eq:sft_gradient}. The accumulation of these interval updates consequently leads to substantially different cumulative update directions between SFT and on-policy paradigms. For cumulative updates, SFT exhibits highly similar directions throughout training~(with cosine similarity close to 1.0), whereas on-policy paradigms exhibit positive yet substantially lower correlation~(with cosine similarity around 0.5). These analyses suggest that, unlike SFT that updates parameters along nearly consistent directions, on-policy paradigms tend to continuously adjust their cumulative parameter update directions throughout training.


\subsection{Applying On-Policy Update Directions to SFT}
Inspired by the continuous efforts to adjust update directions in on-policy paradigms, we investigate whether the resulting cumulative on-policy update direction can serve as an effective optimization principle for transferring their generalization advantage to SFT.

Specifically, we constrain parameter updates in SFT to the directions identified by on-policy paradigms. For consistency with the theoretical analysis above, we represent the model parameters as a $d$-dimensional vector.
Given the parameters before and after on-policy training $(\bm{\theta}_\mathrm{base},\bm{\theta}_{\mathrm{on}})$, we obtain the sign vector $\bm{v}=\textit{sign}(\bm{\theta}_{\mathrm{on}}-\bm{\theta}_\mathrm{base})\in\{-1,0,1\}^{d}$ that determines the direction of the cumulative update and constrains the gradient $\bm{g}\in\mathbb{R}^{d}$ of SFT according to $\bm{v}$ as follows:
\begin{equation}
\label{eq:constrain}
\bm{g}_s = \mathbb{I}\left( \textit{sign}(-\bm{g}) = \bm{v} \right) \odot \bm{g},
\end{equation}
where $\odot$ denotes the Hadamard product, $\textit{sign}(\cdot)$ is the element-wise sign operator, and $\mathbb{I}(\cdot)$ represents the indicator function that retains gradient elements whose signs match $\bm{v}$ while discarding others. 
The constrained gradient $\bm{g}_s$ is then passed to the optimizer for gradient descent. By constraining the gradient at each training step, the parameter updates consistently follow the directions identified by the on-policy paradigm. 
In practice, considering that the regularization terms inherent to the optimizer may affect the imposed constraint, we further constrain the update direction after each optimizer step in Appendix~\ref{app:implementation_details}.
For brevity in the subsequent discussion, we refer to this On-Policy direction-constrained SFT as OPSFT.
By comparing OPSFT with vanilla SFT and the on-policy paradigm that provides the update direction, we can evaluate the ability of the on-policy update direction to transfer the generalization advantage of on-policy paradigms to SFT. 

\subsection{Experimental Results}

\noindent\textbf{Experimental Setups}. 
We train Qwen3-1.7B/4B/8B~\citep{yang2025qwen3} and DeepSeek-R1-Distill-Llama-3-8B~\citep{guo2025deepseek} on the DeepMath~\citep{he2026deepmath}, DAPO~\citep{yu2026dapo}, and Eurus-RL-Code~\citep{cui2025process} datasets with trajectories generated by Qwen3-30B-A3B. For baseline methods, we compare OPSFT with vanilla SFT~\citep{gu2024minillm}, OPD~\citep{lu2025onpolicydistillation}, GRPO~\citep{he2026deepmath}, and OPSFT only sharing the same update locations as on-policy paradigms. For evaluation, we consider both the in-domain~\citep{aime24, aime25, balunovic2026matharena} and out-of-domain datasets~\citep{Zhou2023InstructionFollowingEF, Clark2018ThinkYH, Li2023HaluEvalAL, zellers2019hellaswag, sakaguchi2021winogrande, bisk2020piqa}. 
Due to space limitations, the experiments on OPD are presented in Appendix~\ref{appsec:additional_experiments}.
More details on training and evaluation are provided in Appendix~\ref{app:implementation_details}.

\begin{figure}[tb]
  \centering
  \includegraphics[width=1.0\linewidth]{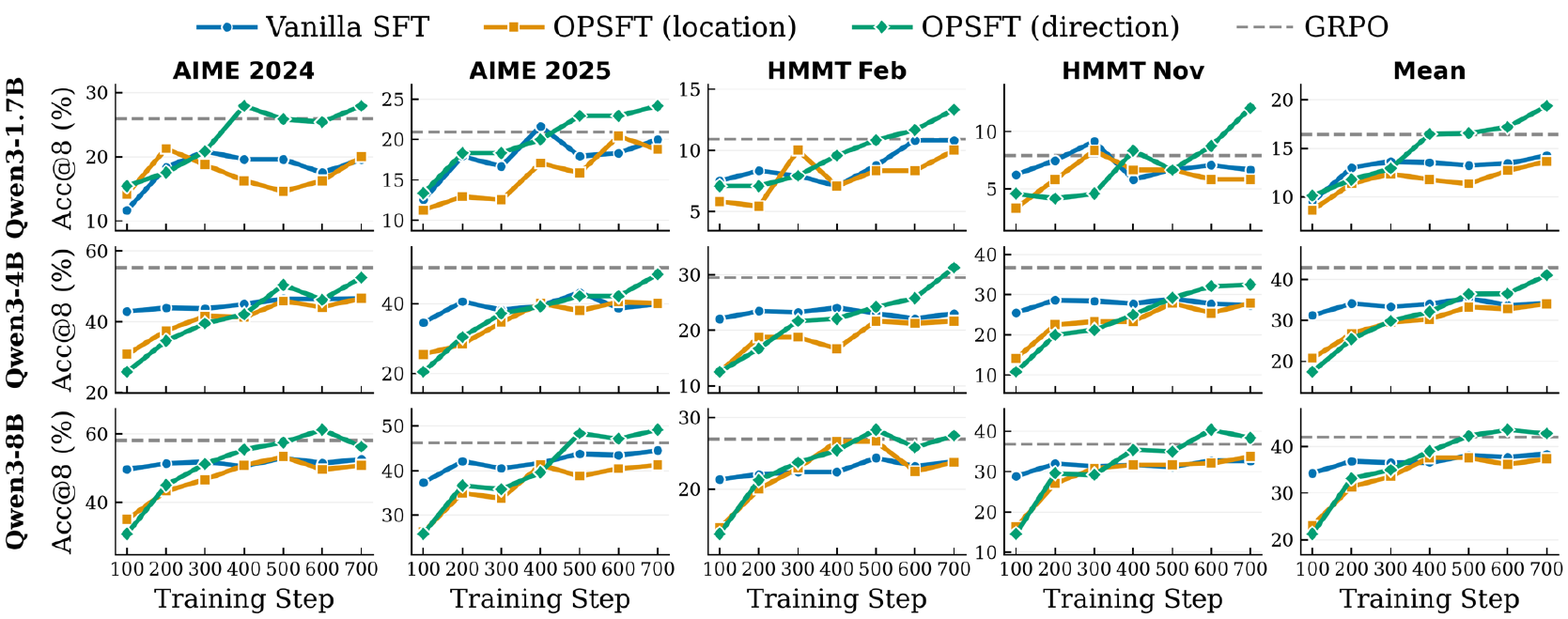}
  \vspace{-0.4cm}
  \caption{Performance comparison across different models and benchmarks. We compare vanilla SFT, OPSFT with location constraints, OPSFT with direction constraints, and the on-policy paradigm that provides update locations and directions~(GRPO).
  }
  \label{fig:generalization_plot}
  \vspace{-0.1cm}
\end{figure}

\begin{table}[t]
\centering
\small
\caption{Performance on out-of-domain benchmarks. We compare the base model, the final checkpoints of vanilla SFT, GRPO, and OPSFT trained on the DeepMath dataset.}
\label{tab:retention}
\setlength\tabcolsep{7pt}
\begin{tabular}{@{}c|c|cccccc|c@{}}
\toprule
Model & Method & IFEval & ARC & HaluEval & Hellaswag & Winogrande & PIQA & Mean \\ \midrule
\multirow{4}{*}{Qwen3-4B} & Base & 87.65 & 87.71 & 72.20 & 52.25 & 65.82 & 74.97 & 73.43 \\
 & SFT & 86.45 & 87.88 & 73.40 & \textbf{52.41} & 66.30 & \textbf{75.46} & 73.65 \\
 & GRPO & 88.61 & 87.88 & 73.70 & 52.25 & 66.22 & 75.03 & 73.95 \\
 & OPSFT & \textbf{88.97} & \textbf{87.97} & \textbf{73.80} & 52.22 & \textbf{66.46} & 75.24 & \textbf{74.11} \\ \midrule
\multirow{4}{*}{Qwen3-8B} & Base & 87.05 & 90.53 & 75.60 & 57.20 & 67.96 & 76.77 & 75.85 \\
 & SFT & 87.53 & 90.61 & 72.70 & \textbf{57.21} & 68.19 & \textbf{76.99} & 75.54 \\
 & GRPO & \textbf{89.09} & 90.53 & 76.00 & 57.14 & 67.80 & 76.66 & 76.20 \\
 & OPSFT & 88.13 & \textbf{90.70} & \textbf{77.10} & 57.15 & \textbf{68.35} & 76.88 & \textbf{76.39} \\ \bottomrule
\end{tabular}
\vspace{-0.3cm}
\end{table}

\noindent\textbf{On-Policy Update Direction Underlies Generalization}. In Figure~\ref{fig:generalization_plot}, we provide the generalization performance of different methods across various benchmarks and model scales.
OPSFT consistently achieves substantial performance gains over vanilla SFT, reaching performance comparable to GRPO that provides the on-policy update directions to OPSFT.
The results indicate that the generalization advantage of on-policy paradigms can be transferred to SFT by sharing the same parameter update direction.
This conclusion challenges the conventional finding that ``SFT memorizes, while RL generalizes"~\citep{chu2025sft}. Once such an update direction is identified, even SFT can achieve strong generalization performance with its updates constrained to this direction.
Furthermore, OPSFT constrained only by the parameter update locations performs even worse than vanilla SFT. This suggests that although prior studies~\citep{mukherjee2026reinforcement,shen2026geometry,yu2026dense} have revealed the distinctive update location of on-policy paradigms, the location alone is insufficient to transfer their generalization advantages. In contrast, extending the constraint from update location $\{0,1\}^d$ to direction $\{-1,0,1\}^d$ achieves substantial gains in generalization performance. These results further highlight the importance of parameter update directions in LLM post-training.

\noindent\textbf{OOD Capability}.
We train the model on the DeepMath dataset and evaluate its performance on datasets from other domains. As shown in Table~\ref{tab:retention}, the results across different benchmarks and model scales demonstrate that OPSFT consistently outperforms vanilla SFT and even slightly surpasses GRPO, which provides the on-policy update direction for OPSFT. This further highlights the effectiveness of on-policy update directions in improving the generalization performance of SFT, consistent with our findings in Figure~\ref{fig:generalization_plot}. The on-policy update direction enables SFT to achieve strong performance on in-domain tasks while facilitating the transfer of newly acquired reasoning capabilities from the math domain to other domains.

\begin{figure}[tb]
  \centering
  \includegraphics[width=1.0\linewidth]{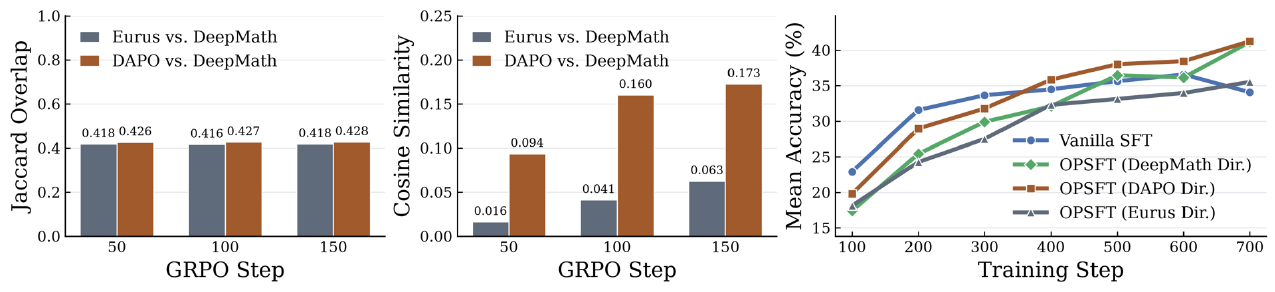}
  \vspace{-0.7cm}
  \caption{We compare parameter update locations~\textit{(left)} and directions~\textit{(middle)} between datasets from the same~(DAPO vs. DeepMath) and different domains~(Eurus vs. DeepMath). Then, we train OPSFT on DeepMath using on-policy update directions identified by different datasets~\textit{(right)}. Mean accuracy is computed across the AIME24/25 and HMMT-Feb/Nov benchmarks.
  }
  \label{fig:cross_dataset}
  \vspace{-0.1cm}
\end{figure}

\noindent\textbf{Reusability Across Datasets and Domains}. 
In Figure~\ref{fig:cross_dataset}, we compare the on-policy update locations and directions across datasets within the same and different domains using Qwen3-4B. First, the update location overlaps are similar between datasets from the same and different domains~(around 0.4), whereas the update directions exhibit substantially higher similarity within the same domain than across different domains. This suggests that update directions play an important role in capturing domain-specific capabilities.
Second, even within the same domain, the update directions identified by different datasets can differ substantially, with the directions identified by DAPO and DeepMath exhibiting a cosine similarity of only around 0.2. This suggests the existence of multiple update directions that can support effective generalization within the same domain. Nevertheless, OPSFT trained on DeepMath using the direction identified by DAPO achieves strong generalization, even slightly outperforming OPSFT using the direction identified by DeepMath itself. This result indicates that on-policy update directions can be reused across datasets within the same domain. In contrast, such reusability does not extend across domains, as evidenced by the weak performance of OPSFT on the math domain when using the direction identified by Eurus, a code-domain dataset.

According to the above experiments, the on-policy update direction enables SFT to achieve strong generalization on in-domain and out-of-domain tasks with reusability across datasets within the same domain. These findings indicate that the on-policy update direction underlies strong generalization in LLM post-training and reveal its potential to improve the post-training pipelines.
\begin{table}[t]
\centering
\small
\caption{Accuracy and training time of different methods across multiple models and benchmarks. All models are trained on the DeepMath dataset. The time of OPSFT includes both identifying the on-policy update direction and the subsequent OPSFT training.}
\setlength\tabcolsep{6.5pt}
\begin{tabular}{@{}c|cc|cccc|c@{}}
\toprule
Model & Method & Time & AIME24 & AIME25 & HMMT25-Feb & HMMT25-Nov & Mean \\ \midrule
\multirow{4}{*}{Qwen3-1.7B} & GRPO & 4.9h & 16.67 & 17.08 & 7.92 & 6.67 & 12.09 \\
 & SFT & 3.5h & 19.48 & 17.92 & 8.96 & 6.25 & 13.15 \\
 & DFT & 3.9h & 18.25 & 17.72 & \textbf{10.00} & 6.67 & 13.16 \\
 & OPSFT & \textbf{2.3h} & \textbf{20.00} & \textbf{21.67} & 9.58 & \textbf{9.17} & \textbf{15.11} \\ \midrule
\multirow{4}{*}{Qwen3-4B} & GRPO & 16.5h & 50.83 & 45.31 & \textbf{26.35} & \textbf{33.33} & 38.96 \\
 & SFT & 8.1h & 46.67 & 39.79 & 23.02 & 27.40 & 34.22 \\
 & DFT & 8.6h & 46.44 & 39.19 & 25.08 & 28.04 & 34.69 \\
 & OPSFT & \textbf{6.4h} & \textbf{52.92} & \textbf{50.42} & 25.83 & 31.25 & \textbf{40.11} \\ \midrule
\multirow{4}{*}{Qwen3-8B} & GRPO & 19.3h & 54.17 & \textbf{44.58} & \textbf{28.33} & 34.17 & 40.31 \\
 & SFT & 13.2h & 52.60 & 44.48 & 23.85 & 32.71 & 38.41 \\
 & DFT & 13.9h & 54.23 & 43.92 & 24.50 & 33.23 & 38.97 \\
 & OPSFT & \textbf{8.9h} & \textbf{57.92} & 42.92 & 26.67 & \textbf{39.17} & \textbf{41.67} \\ \midrule
\multirow{4}{*}{\begin{tabular}[c]{@{}c@{}}DeepSeek-R1-\\ LLaMA-8B\end{tabular}} & GRPO & 18.2h & 36.25 & 29.17 & 17.50 & 17.92 & 25.21 \\
 & SFT & 13.6h & 24.16 & 23.75 & 12.50 & 11.67 & 18.02 \\
 & DFT & 14.1h & 31.67 & 24.58 & 14.58 & 8.75 & 19.90 \\
 & OPSFT & \textbf{8.1h} & \textbf{37.91} & \textbf{30.41} & \textbf{17.92} & \textbf{20.00} & \textbf{26.56} \\ \bottomrule
\end{tabular}
\label{tab:application_1}
\vspace{-0.0cm}
\end{table}
\begin{table}[t]
\centering
\small
\caption{Performance comparisons between SFT and OPSFT across multiple benchmarks using different models post-trained by GRPO. All models are trained on the DeepMath dataset.}
\begin{tabular}{@{}c|c|cccc|c@{}}
\toprule
Model & Method & AIME24 & AIME25 & HMMT25-Feb & HMMT25-Nov & Mean \\ \midrule
\multirow{3}{*}{Qwen3-1.7B} & Base (post-trained) & 16.67 & 17.08 & 7.92 & 6.67 & 12.09 \\
 & SFT & 17.92 & 19.17 & \textbf{10.83} & 6.25 & 13.54 \\
 & OPSFT & \textbf{26.25} & \textbf{22.08} & 10.42 & \textbf{8.33} & \textbf{16.77} \\ \midrule
\multirow{3}{*}{Qwen3-4B} & Base (post-trained) & 50.83 & \textbf{45.31} & 26.35 & 33.33 & 38.96 \\
 & SFT & 50.00 & 42.08 & 24.17 & 28.75 & 36.25 \\
 & OPSFT & \textbf{56.67} & 44.17 & \textbf{27.92} & \textbf{36.67} & \textbf{41.36} \\ \midrule
\multirow{3}{*}{Qwen3-8B} & Base (post-trained) & 54.17 & 44.58 & \textbf{28.33} & 34.17 & 40.31 \\
 & SFT & 52.92 & 39.58 & 25.00 & 32.50 & 37.50 \\
 & OPSFT & \textbf{57.08} & \textbf{48.75} & 27.91 & \textbf{35.83} & \textbf{42.39} \\ \midrule
\multirow{3}{*}{\begin{tabular}[c]{@{}c@{}}DeepSeek-R1-\\ LLaMA-8B\end{tabular}} & Base (post-trained) & 36.25 & 29.17 & 17.50 & \textbf{17.92} & 25.21 \\
 & SFT & 21.25 & 22.08 & 11.67 & 10.42 & 16.36 \\
 & OPSFT & \textbf{37.10} & \textbf{31.67} & \textbf{17.92} & \textbf{17.92} & \textbf{26.15} \\ \bottomrule
\end{tabular}
\label{tab:application_2}
\vspace{-0.1cm}
\end{table}
\begin{table}[t]
\small
\begin{minipage}[c]{.49\textwidth}
\centering
\caption{Accuracy and training time comparisons on code tasks. Experiments are conducted with Qwen3-4B on the Eurus dataset.}
\label{tab: application_code_1}
\setlength\tabcolsep{3pt}
  \begin{tabular}{@{}cc|ccc|c@{}}
\toprule
Method & Time & HumanEval+ & MBPP+ & LCBv6 & Mean \\ \midrule
GRPO & 20.3h & 80.79 & \textbf{69.04} & 21.75 & 57.19 \\
SFT & 9.8h & 80.50 & 63.90 & 18.51 & 54.30 \\
OPSFT & \textbf{9.5h} & \textbf{84.00} & 67.30 & \textbf{22.90} & \textbf{58.07} \\ \bottomrule
\end{tabular}
\end{minipage}
\hfill
\begin{minipage}[c]{.47\textwidth}%
\centering
\caption{Performance comparisons between SFT and OPSFT using Qwen3-4B post-trained by GRPO on the Eurus dataset.}
\label{tab: application_code_2}
\setlength\tabcolsep{3pt}
\begin{tabular}{@{}c|ccc|c@{}}
\toprule
Method & HumanEval+ & MBPP+ & LCBv6 & Mean \\ \midrule
Base & 80.79 & \textbf{69.04} & 21.75 & 57.19 \\
SFT & 70.60 & 50.30 & 16.22 & 48.71 \\
OPSFT & \textbf{85.80} & 68.20 & \textbf{26.15} & \textbf{60.05} \\ \bottomrule
\end{tabular}
	\end{minipage}
\vspace{-0.1cm}
\end{table}
\begin{table}[t]
\small
\centering
\caption{Ablation studies of the parameter precisions. We report the accuracy and the proportion of updated parameters across different methods. Experiments are conducted with Qwen3-8B. }
\setlength\tabcolsep{5.3pt}
\begin{tabular}{@{}c|cc|cccc|c@{}}
\toprule
Precision & Method & Updated Param. & AIME24 & AIME25 & HMMT25-Feb & HMMT25-Nov & Mean \\ \midrule
\multirow{2}{*}{FP32} & Base & 0.000\% & 27.08 & 20.83 & 9.17 & 8.33 & 16.35 \\
 & GRPO & 9.499\% & 58.13 & 46.15 & 26.98 & 36.77 & 42.01 \\ \midrule
\multirow{2}{*}{BF16} & SFT & 2.702\% & 54.16 & 42.50 & 27.08 & 29.58 & 38.33 \\
 & OPSFT & 0.408\% & 56.77 & 45.73 & 26.15 & 35.10 & \textbf{40.94} \\ \midrule
\multirow{2}{*}{FP32} & SFT & 89.286\% & 52.60 & 44.48 & 23.85 & 32.71 & 38.41 \\
 & OPSFT & 9.454\% & 56.25 & 49.17 & 27.50 & 38.33 & \textbf{42.81} \\ \bottomrule
\end{tabular}
\label{tab:abl_precision}
\vspace{-0.1cm}
\end{table}

\begin{figure}[t]
  \centering
  \includegraphics[width=1.0\linewidth]{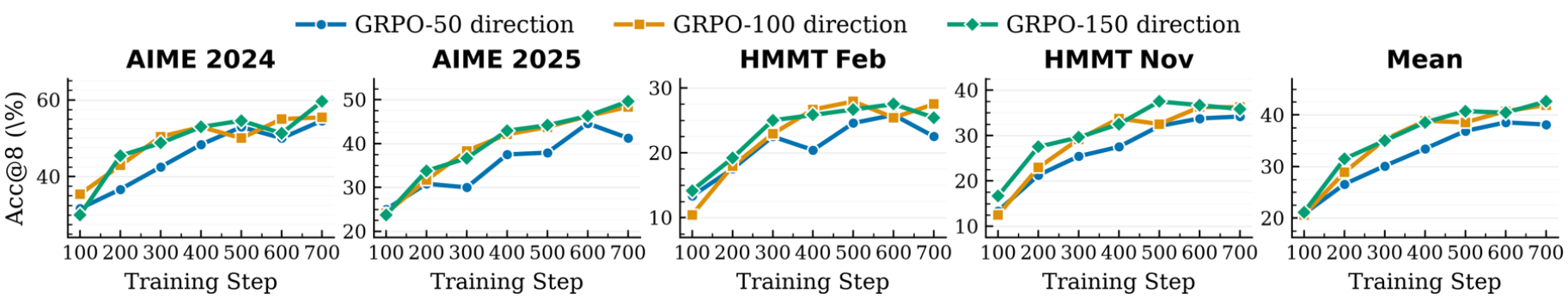}
  \vspace{-0.6cm}
  \caption{Accuracy evolution of OPSFT using the update direction identified at different GRPO training steps. Experiments are conducted using Qwen3-8B on the DeepMath dataset.
  }
  \label{fig:abl_subspace}
  \vspace{-0.1cm}
\end{figure}

\begin{figure}[t]
  \centering
  \includegraphics[width=1.0\linewidth]{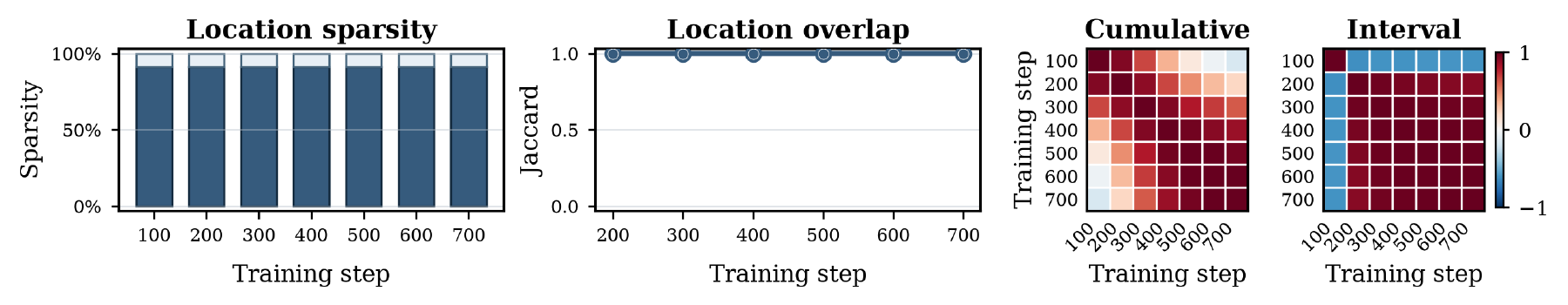}
  \vspace{-0.4cm}
  \caption{Update behaviors of OPSFT when training Qwen3-8B on the DeepMath dataset. We report the location sparsity, Jaccard overlap of update locations, and matrix-level cosine similarity of cumulative and interval updates across different training steps.
  }
  \label{fig:trajectory}
  \vspace{-0.1cm}
\end{figure}

\begin{figure}[t]
  \centering
  \includegraphics[width=1.0\linewidth]{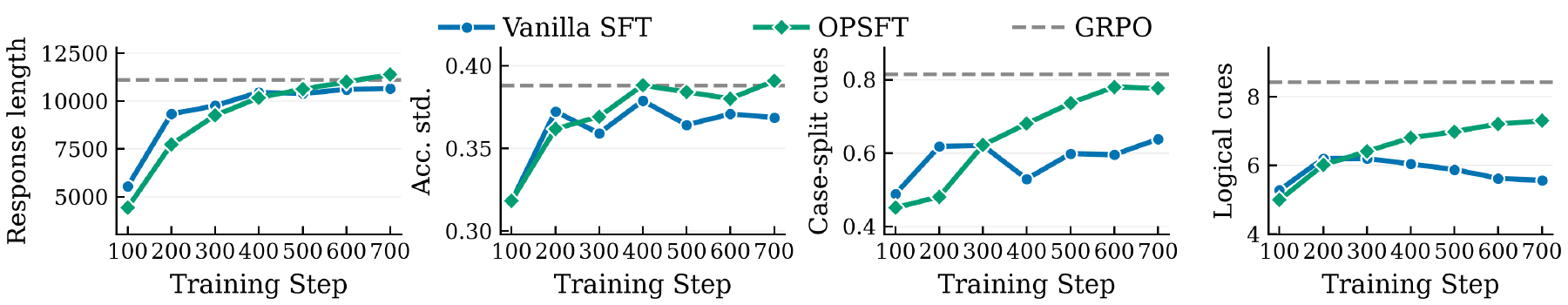}
  \vspace{-0.6cm}
  \caption{Reasoning behavior comparisons. We report response length, accuracy standard deviation, the number of case-split cues per 1k words, and the number of logical cues per 1k words throughout training with Qwen3-8B on the DeepMath dataset.
  }
  \label{fig:reasoning_behavior}
  \vspace{-0.1cm}
\end{figure}

\section{Applications of On-Policy Update Direction}
Motivated by the ability of the on-policy update direction to support strong generalization, we further leverage this direction to combine the generalization advantage of on-policy paradigms with the advantages of SFT. First, we perform a small number of GRPO steps to identify an update direction that supports strong generalization, and then apply OPSFT to achieve efficient training~(Section~\ref{sec:application_1}). Second, given a model post-trained by an on-policy paradigm, we conduct OPSFT along its original update direction, thereby leveraging newly acquired high-quality trajectories to further improve the model without disrupting the capabilities learned during on-policy training~(Section~\ref{sec:application_2}). Training and evaluation details are provided in Appendix~\ref{app:implementation_details}.

\subsection{Training Efficiency Improvement}
\label{sec:application_1}
We train the model on the DeepMath dataset using GRPO~(100 steps), SFT~(700 steps), DFT~\citep{wu2026generalization}~(700 steps), and OPSFT~(50 GRPO steps for direction identification followed by 100 OPSFT steps). The results across different model scales, architectures, and benchmarks are reported in Table~\ref{tab:application_1}.
First, compared with SFT and its improved version DFT, OPSFT achieves substantially better performance while requiring less training time by leveraging the on-policy update direction. For example, on DeepSeek-R1-Distill-LLaMA-8B, OPSFT achieves a mean accuracy of 26.56 in only 8.1h, substantially outperforming DFT that achieves 19.90 with 14.1h of training.
Second, compared with GRPO, OPSFT reduces training time by more than 50\%~(\textit{e.g.}, 8.9h vs. 19.3h on Qwen3-8B) while achieving better generalization performance~(\textit{e.g.}, 41.67 vs. 40.31 mean accuracy on Qwen3-8B).
These results demonstrate that update directions identified during the early stages of on-policy training are already sufficient to substantially improve the generalization of SFT. More importantly, OPSFT is not limited by the on-policy paradigm used to identify the direction, as it can even outperform 100-step GRPO when using an update direction identified after only 50 GRPO steps.
Moreover, we provide results on code tasks in Table~\ref{tab: application_code_1}. The results are consistent with those on math tasks, demonstrating the generality of our findings across different domains.

\subsection{Improve Models After On-Policy Post-Training}
\label{sec:application_2}
In Table~\ref{tab:application_2}, we investigate whether OPSFT can further improve post-trained models by leveraging newly acquired high-quality trajectories. We first conduct 100 steps of GRPO and then continue training the resulting model on reasoning trajectories from the DeepMath dataset for 700 steps. Directly applying SFT to the post-trained model leads to performance degradation~(\textit{e.g.}, the mean accuracy of Qwen3-4B drops from 38.96 to 36.25). This suggests that even high-quality trajectories cannot be directly used to improve a post-trained model, as SFT may disrupt the capabilities acquired during on-policy post-training. In contrast, by constraining parameter updates to the on-policy update direction, OPSFT further improves the post-trained model~(\textit{e.g.}, increasing the mean accuracy of Qwen3-4B from 38.96 to 41.36).
Furthermore, we provide results on code tasks in Table~\ref{tab: application_code_2}, which are consistent with those observed on math tasks. As high-quality trajectories can continuously accumulate in practice, OPSFT provides a promising way for incrementally improving post-trained models without restarting the entire post-training process from the base model.

\subsection{Ablation Studies and Analysis}

\noindent\textbf{Effects of the Direction from Different Training Steps}.
We train OPSFT using update directions identified by the on-policy paradigm at different training stages. The performance across different benchmarks is reported in Figure~\ref{fig:abl_subspace}.
First, we observe that update directions identified at later stages of GRPO consistently lead to better generalization performance in OPSFT. Moreover, the marginal benefit of using later-stage update directions gradually diminishes as GRPO training progresses. Specifically, the improvement from the direction identified at 100 steps over that identified at 50 steps is larger than the improvement from the 150-step direction over the 100-step direction.

\noindent\textbf{Effects of Parameter Precisions}. In Table~\ref{tab:abl_precision}, we report the performance across different models and benchmarks under BF16 and FP32 parameter precision. Under BF16 precision, many small gradients result in parameter updates below the numerical precision of BF16 and are therefore rounded to zero, resulting in relatively sparse updates for vanilla SFT~(only 2.702\% of parameters are updated). Nevertheless, OPSFT induces substantially sparser updates than vanilla SFT by constraining parameter updates to the on-policy update direction. Despite its extremely sparse updates, OPSFT still significantly outperforms vanilla SFT under BF16 precision~(40.94\% vs. 38.33\%), achieving a 24.59\% improvement over the base model while updating only 0.408\% of the parameters.
Under FP32, the updates of both SFT variants become substantially denser. Vanilla SFT updates approximately 90\% of the parameters, whereas OPSFT updates only 9.45\%, comparable to GRPO. Moreover, FP32 precision leads to better performance than BF16 precision, with OPSFT even outperforming GRPO, which is used to identify its update direction~(42.81\% vs. 42.01\%).

\noindent\textbf{Optimization Behavior of OPSFT}. In Figure~\ref{fig:trajectory}, we analyze the evolution of update locations and matrix-level directions in OPSFT throughout training. 
For update locations, OPSFT updates are constrained to a subset of parameters by the on-policy update direction, resulting in dense updates within this subset. For update directions, we find that OPSFT exhibits substantial differences between the early and later stages of training. This behavior suggests that the parameter-level direction~(\textit{i.e.}, sign) constraint of OPSFT redirects the SFT optimization trajectory toward a matrix-level direction conducive to generalization, after which OPSFT can continue optimizing along this direction.

\noindent\textbf{Reasoning Behavior}. In Figure~\ref{fig:reasoning_behavior}, we compare vanilla SFT and OPSFT with GRPO in terms of reasoning behaviors. First, we find that vanilla SFT, OPSFT, and GRPO exhibit similar response lengths. Moreover, compared with vanilla SFT, OPSFT exhibits a higher standard deviation in accuracy across multiple responses, approaching that of GRPO. This suggests that incorporating the on-policy update direction promotes greater uncertainty in the model's reasoning outcomes. To investigate these behaviors at a finer granularity, we analyze the frequencies of case-splitting cues~(\textit{e.g.}, \texttt{if}, \texttt{otherwise}, \texttt{consider a case}) and logical cues~(\textit{e.g.}, \texttt{therefore}, \texttt{because}) throughout training. The results indicate that the direction constraint makes the reasoning behaviors of OPSFT more closely resemble those of on-policy paradigms, rather than simply inducing longer reasoning.

\section{Conclusion}
In this paper, we explore whether there exists a parameter update behavior in on-policy paradigms that can be leveraged to transfer their generalization advantage to SFT. First, our theoretical and experimental analyses reveal that on-policy paradigms tend to continuously adjust their parameter update directions during training. Inspired by this phenomenon, we propose OPSFT, which constrains parameter updates of SFT to the direction identified by the on-policy paradigm. Extensive experiments show that OPSFT achieves performance comparable to that of the corresponding on-policy paradigm, demonstrating that the on-policy update direction can serve as an effective optimization principle for improving the generalization of SFT. Based on this finding, OPSFT enables integrating the strong generalization of on-policy paradigms with the advantages of SFT, including efficient training and the ability to leverage high-quality trajectories. For future work, we will investigate how different components of on-policy paradigms contribute to identifying generalization-friendly directions, and how the property of reasoning trajectories affects model generalization through OPSFT. For limitation discussions, please refer to Appendix~\ref{app:limitations}.

\clearpage

\bibliography{iclr2027_conference}
\bibliographystyle{iclr2027_conference}

\clearpage

\renewcommand{\thefigure}{A\arabic{figure}}
\renewcommand{\thetable}{A\arabic{table}}
\renewcommand{\theequation}{A\arabic{equation}}

\appendix

\section{Implementation Details}
\label{app:implementation_details}
\subsection{Training Datasets}
\noindent\textbf{On-policy paradigms}. We filter the DeepMath~\citep{he2026deepmath} dataset to select 57K samples with a difficulty level
greater than or equal to 6 to form the math RL data, and use Eurus-RL-Code~\citep{cui2025process} as the code RL data, which consists of 25K samples.  

\noindent\textbf{SFT}. For math, we construct SFT trajectories from the DeepMath~\citep{he2026deepmath} training prompts using \texttt{Qwen3-30B-A3B-Instruct-2507}. We decode one completion per prompt with temperature \(0.6\), top-\(p=0.95\), and a maximum of \(16{,}384\) newly generated tokens (with an \(18{,}432\)-token model context limit). A trajectory is retained only when the teacher response is judged correct against the instance ground truth by the mathematical answer verifier (\texttt{math\_verify}); malformed, empty, or incorrect generations are discarded. 
For code, we use the same model and protocol following the math data-generation setting with Eurus~\citep{cui2025process}. We sample one completion per prompt and retain it only if the evaluator successfully executes the generated program against the reference tests. We require every response to contain a Python code block and store execution metadata together with the prompt, response, and ground truth. After verification, the math dataset contains \(44{,}810\) training and \(914\) examples. The final Eurus code dataset contains \(9{,}585\) training examples.

\subsection{Training Settings}
\noindent\textbf{On-policy paradigms}.
For on-policy paradigms in the main paper, we apply Group Relative Policy Optimization~(GRPO)~\citep{shao2024deepseekmath}. A reward of 1.0 is given when the final answer is correct in math reasoning or when all unit tests pass in code generation; otherwise, the reward is 0.0. The training hyperparameters in math and code training are put in Table~\ref{apptab:math RL hyper-parameters} and Table~\ref{apptab:code RL hyper-parameters}, respectively.

\begin{table}[t]  
\begin{center}
\begin{minipage}[t]{0.48\linewidth}
\caption{Training hyperparameters of GRPO in math tasks.}
\label{apptab:math RL hyper-parameters}
\centering
\begin{tabular}{ll}
\toprule
Hyper-parameter & Value \\
\midrule
Train Batch Size & 128  \\
Micro Batch Size & 128 \\
Rollout $n$ & 8 \\
Maximum Prompt Length & 2048 \\
Maximum Response Length & 16,384 \\
Temperature & 1.0 \\
Top-p & 1.0 \\
LR & $1\times 10^{-6}$ \\
Optimization Steps & 150 \\
KL Coefficient & 0.0 \\
\bottomrule
\end{tabular}
\end{minipage}  
\hfil
\begin{minipage}[t]{0.48\linewidth}
\caption{Training hyperparameters of GRPO in code tasks.}
\label{apptab:code RL hyper-parameters}
\centering
\begin{tabular}{ll}
\toprule
Hyper-parameter & Value \\
\midrule
Train Batch Size & 128  \\
Micro Batch Size & 128 \\
Rollout $n$ & 8 \\
Maximum Prompt Length & 2048 \\
Maximum Response Length & 8192 \\
Temperature & 1.0 \\
Top-p & 1.0 \\
LR & $1\times 10^{-6}$ \\
Optimization Steps & 150 \\
KL Coefficient & 0.0 \\
\bottomrule
\end{tabular}
\end{minipage} 
\end{center}
\end{table}

\noindent\textbf{SFT}.
For both math and code experiments, we perform SFT and OPSFT using the cross-entropy objective. All experiments use FP32 training with FSDP2 sharding over 8 NVIDIA H20 GPUs. We use a global batch size of 64, a maximum sequence length of 16,384 tokens, and right truncation for overlength examples. Gradient checkpointing is enabled. Optimization is performed with AdamW ($\beta_1=0.9$, $\beta_2=0.95$), weight decay of $0.01$, and gradient-norm clipping at $1.0$. We use a constant learning-rate schedule with no warm-up steps. Unless otherwise specified, the learning rate is $1\times10^{-7}$, and the random seed is fixed to 42. Models are trained for 700 optimizer steps, and checkpoints are saved every 100 steps.
For the experiment in Table~\ref{sec:application_1} and Table~\ref{tab: application_code_1}, we retain all settings above and use a learning rate of $1\times10^{-6}$.
Unless otherwise specified, all experiments use FP32 parameter precision by default.
Moreover, OPSFT enforces the direction constraint at both the gradient and parameter levels. At each optimization step, gradients that induce updates opposite to the cumulative GRPO update direction are masked out. After the optimizer step, we further restore any parameter whose cumulative displacement from the initial weights has moved in the opposite direction. This two-level constraint ensures that neither AdamW's weight decay nor its momentum-based updates can violate the prescribed on-policy update direction.

\subsection{Evaluation Details}
For the evaluation of math reasoning, we select four competition-level benchmarks: AIME24~\citep{aime24}, AIME25~\citep{aime25}, HMMT25~(February)~\citep{balunovic2026matharena}, and HMMT25~(November)~\citep{balunovic2026matharena}. For the evaluation of code generation, we select three test sets: HumanEval+, MBPP+~\citep{liu2023your}, and LiveCodeBench (v6 only, February 2025$\sim$May 2025)~\citep{jain2025livecodebench}. In all
evaluations, we set the temperature to 1.0, top-p to 1.0, and the maximum generation length to 16,384. On each math reasoning benchmark, we sample 8 solutions for each problem, whereas on each code generation benchmark, we sample 4 solutions per problem. We then report the average accuracy of each model on each benchmark. We
adopt Math-Verify as a verifier to validate answer correctness for math reasoning benchmarks.

\subsection{Pseudo Code of OPSFT}
\begin{lstlisting}[
    style=pseudocode,
    caption={Pseudo code of OPSFT.
    $M$ is the parameter update location extracted from on-policy paradigms, and $D$ is the
    element-wise parameter update direction.},
    label={lst:opd_constrained_sft_step}
]
# Inputs:
#   theta_0[name]: parameter value before SFT
#   M[name]: exact-support boolean mask
#   D[name]: cumulative OPD/GRPO parameter update
#   batch: one SFT minibatch
#   optimizer: e.g., AdamW
optimizer.zero_grad()
# Compute the standard SFT gradient on one minibatch.
loss = compute_sft_loss(model, batch)
loss.backward()
# Stage 1: constrain the instantaneous SFT gradient.
for name, parameter in model.named_parameters():
    if parameter.grad is None:
        continue
    gradient = parameter.grad
    # Location: only OPD exact-support coordinates receive gradients.
    gradient[~M[name]] = 0
    # Direction: -gradient is the first-order parameter movement.
    # Block gradients whose induced movement opposes D[name].
    opposite_direction = (
        M[name]
        & (gradient != 0)
        & (D[name] != 0)
        & (gradient * D[name] > 0)
    )
    gradient[opposite_direction] = 0
clip_gradient_norm(model.parameters(), max_norm)
optimizer.step()
# Stage 2: hard projection after the optimizer update.
with no_gradient():
    for name, parameter in model.named_parameters():
        # Enforce exact support.
        parameter[~M[name]] = theta_0[name][~M[name]]
        # Enforce agreement between cumulative SFT displacement and D[name].
        displacement = parameter - theta_0[name]
        wrong_direction = (
            M[name]
            & (D[name] != 0)
            & (displacement * D[name] < 0)
        )
        parameter[wrong_direction] = theta_0[name][wrong_direction]
\end{lstlisting}

\section{Additional Experiments}
\label{appsec:additional_experiments}
\noindent\textbf{Experiments on other on-policy training paradigms}.
In Table~\ref{apptab:opd}, we conduct experiments on the OPD update direction following the settings in Figure~\ref{fig:generalization_plot}. For OPD, we use Qwen3-30B-A3B as the teacher model and Qwen3-1.7B as the student model, and train the student for 150 steps. All other training settings follow G-OPD~\citep{yang2026learning}. We observe consistent results with our previous experiments. Compared with vanilla SFT, OPSFT with both location and direction constraints on the OPD update direction achieves substantial performance improvements~(16.71\% vs. 13.83\% of mean accuracy), which are similar to the OPD performance~(16.71\% vs. 16.75\% of mean accuracy). These results demonstrate the generality of our findings across different on-policy paradigms.

\begin{table}[t]
\centering
\small
\caption{Accuracy of different methods on Qwen3-1.7B using the DeepMath dataset.}
\label{apptab:opd}
\begin{tabular}{@{}c|cccc|c@{}}
\toprule
Method & AIME24 & AIME25 & HMMT25-Feb & HMMT25-Nov & Mean \\ \midrule
Base & 13.54 & 11.35 & 4.79 & 4.68 & 8.59 \\
OPD & 24.17 & 20.93 & 11.67 & 10.21 & 16.75 \\
Vanilla SFT & 18.90 & 18.60 & 10.40 & 7.40 & 13.83 \\
OPSFT (location) & 18.90 & 18.90 & 9.40 & 6.70 & 13.48 \\
OPSFT (direction) & 23.02 & 23.95 & 10.93 & 8.95 & 16.71 \\ \bottomrule
\end{tabular}
\end{table}

\noindent\textbf{Raw data of Figure~\ref{fig:generalization_plot}}. We provide the raw data of Qwen3-1.7B, Qwen3-4B, and Qwen3-8B in Table~\ref{apptab:qwen3-1p7b-trajectory}, Table~\ref{apptab:qwen3-4b-trajectory}, and Table~\ref{apptab:qwen3-8b-trajectory}, respectively.

\begin{table}[t]
\centering
\caption{Training trajectories for Qwen3-1.7B. Entries report Acc@8 (\%).}
\label{apptab:qwen3-1p7b-trajectory}
\small
\setlength{\tabcolsep}{4.5pt}
\begin{tabular}{llrrrrr}
\toprule
Method & Step & AIME24 & AIME25 & HMMT Feb & HMMT Nov & Mean \\
\midrule
\multirow{7}{*}{Vanilla SFT}
 & 100 & 11.67 & 12.50 & 7.50 & 6.25 & 9.48 \\
 & 200 & 18.33 & 17.92 & 8.33 & 7.50 & 13.02 \\
 & 300 & 20.83 & 16.67 & 7.92 & 9.17 & 13.65 \\
 & 400 & 19.58 & 21.67 & 7.08 & 5.83 & 13.54 \\
 & 500 & 19.58 & 17.92 & 8.75 & 6.67 & 13.23 \\
 & 600 & 17.50 & 18.33 & 10.83 & 7.08 & 13.44 \\
 & 700 & 19.58 & 20.00 & 10.83 & 6.67 & 14.27 \\
\midrule
\multirow{7}{*}{OPSFT (location)}
 & 100 & 14.17 & 11.25 & 5.83 & 3.33 & 8.65 \\
 & 200 & 21.25 & 12.92 & 5.42 & 5.83 & 11.35 \\
 & 300 & 18.75 & 12.50 & 10.00 & 8.33 & 12.40 \\
 & 400 & 16.25 & 17.08 & 7.08 & 6.67 & 11.77 \\
 & 500 & 14.58 & 15.83 & 8.33 & 6.67 & 11.35 \\
 & 600 & 16.25 & 20.42 & 8.33 & 5.83 & 12.71 \\
 & 700 & 20.00 & 18.75 & 10.00 & 5.83 & 13.65 \\
\midrule
\multirow{7}{*}{OPSFT (direction)}
 & 100 & 15.42 & 13.33 & 7.08 & 4.58 & 10.10 \\
 & 200 & 17.50 & 18.33 & 7.08 & 4.17 & 11.77 \\
 & 300 & 20.83 & 18.33 & 7.92 & 4.58 & 12.92 \\
 & 400 & 27.92 & 20.00 & 9.58 & 8.33 & 16.46 \\
 & 500 & 25.83 & 22.92 & 10.83 & 6.67 & 16.56 \\
 & 600 & 25.42 & 22.92 & 11.67 & 8.75 & 17.19 \\
 & 700 & 27.92 & 24.17 & 13.33 & 12.08 & 19.38 \\
\bottomrule
\end{tabular}
\end{table}

\begin{table}[t]
\centering
\caption{Training trajectories for Qwen3-4B. Entries report Acc@8 (\%).}
\label{apptab:qwen3-4b-trajectory}
\small
\setlength{\tabcolsep}{4.5pt}
\begin{tabular}{llrrrrr}
\toprule
Method & Step & AIME24 & AIME25 & HMMT Feb & HMMT Nov & Mean \\
\midrule
\multirow{7}{*}{Vanilla SFT}
 & 100 & 42.92 & 34.48 & 22.08 & 25.52 & 31.25 \\
 & 200 & 43.96 & 40.52 & 23.44 & 28.65 & 34.14 \\
 & 300 & 43.75 & 38.12 & 23.23 & 28.44 & 33.38 \\
 & 400 & 45.00 & 39.27 & 24.06 & 27.71 & 34.01 \\
 & 500 & 46.56 & 43.12 & 23.02 & 29.06 & 35.44 \\
 & 600 & 46.46 & 38.65 & 22.08 & 27.71 & 33.73 \\
 & 700 & 46.67 & 39.79 & 23.02 & 27.40 & 34.22 \\
\midrule
\multirow{7}{*}{OPSFT (location)}
 & 100 & 30.83 & 25.42 & 12.50 & 14.17 & 20.73 \\
 & 200 & 37.50 & 28.33 & 18.75 & 22.50 & 26.77 \\
 & 300 & 41.67 & 34.58 & 18.75 & 23.33 & 29.58 \\
 & 400 & 41.25 & 40.00 & 16.67 & 23.33 & 30.31 \\
 & 500 & 45.83 & 37.92 & 21.67 & 27.92 & 33.34 \\
 & 600 & 44.17 & 40.42 & 21.25 & 25.42 & 32.81 \\
 & 700 & 46.67 & 40.00 & 21.67 & 27.92 & 34.06 \\
\midrule
\multirow{7}{*}{OPSFT (location)}
 & 100 & 25.83 & 20.42 & 12.50 & 10.83 & 17.39 \\
 & 200 & 34.58 & 30.42 & 16.67 & 20.00 & 25.42 \\
 & 300 & 39.58 & 37.08 & 21.67 & 21.25 & 29.89 \\
 & 400 & 42.08 & 39.17 & 22.08 & 25.00 & 32.08 \\
 & 500 & 50.42 & 42.08 & 24.17 & 29.17 & 36.46 \\
 & 600 & 46.25 & 42.08 & 25.80 & 32.08 & 36.55 \\
 & 700 & 52.50 & 48.33 & 31.25 & 32.50 & 41.14 \\
\bottomrule
\end{tabular}
\end{table}

\begin{table}[t]
\centering
\caption{Training trajectories for Qwen3-8B. Entries report Acc@8 (\%).}
\label{apptab:qwen3-8b-trajectory}
\small
\setlength{\tabcolsep}{4.5pt}
\begin{tabular}{llrrrrr}
\toprule
Method & Step & AIME24 & AIME25 & HMMT Feb & HMMT Nov & Mean \\
\midrule
\multirow{7}{*}{Vanilla SFT}
 & 100 & 49.69 & 37.29 & 21.35 & 28.85 & 34.30 \\
 & 200 & 51.35 & 42.08 & 22.08 & 31.98 & 36.87 \\
 & 300 & 51.88 & 40.52 & 22.40 & 31.35 & 36.54 \\
 & 400 & 50.52 & 41.67 & 22.40 & 31.67 & 36.56 \\
 & 500 & 53.02 & 43.75 & 24.38 & 31.15 & 38.08 \\
 & 600 & 51.46 & 43.44 & 23.23 & 32.71 & 37.71 \\
 & 700 & 52.60 & 44.48 & 23.85 & 32.71 & 38.41 \\
\midrule
\multirow{7}{*}{OPSFT (location)}
 & 100 & 35.00 & 26.25 & 14.58 & 16.25 & 23.02 \\
 & 200 & 43.33 & 35.00 & 20.00 & 27.08 & 31.35 \\
 & 300 & 46.67 & 33.75 & 22.92 & 30.83 & 33.54 \\
 & 400 & 50.83 & 41.25 & 26.67 & 31.67 & 37.61 \\
 & 500 & 53.33 & 38.75 & 26.67 & 31.67 & 37.61 \\
 & 600 & 49.58 & 40.42 & 22.50 & 32.08 & 36.14 \\
 & 700 & 50.83 & 41.25 & 23.75 & 33.75 & 37.39 \\
\midrule
\multirow{7}{*}{OPSFT (direction)}
 & 100 & 30.83 & 25.83 & 13.75 & 14.58 & 21.25 \\
 & 200 & 45.00 & 36.67 & 21.25 & 29.58 & 33.12 \\
 & 300 & 51.25 & 35.83 & 23.75 & 29.17 & 35.00 \\
 & 400 & 55.42 & 39.58 & 25.42 & 35.42 & 38.96 \\
 & 500 & 57.50 & 48.33 & 28.33 & 35.00 & 42.29 \\
 & 600 & 61.25 & 47.08 & 25.83 & 40.42 & 43.64 \\
 & 700 & 56.25 & 49.17 & 27.50 & 38.33 & 42.81 \\
\bottomrule
\end{tabular}
\end{table}

\noindent\textbf{Ablations of Update Location and Direction}.
In Table~\ref{apptab:random}, we provide ablation studies of the update location and direction identified by the on-policy paradigm. First, compared with randomly constraining the parameter update locations, selecting the trainable parameters following the on-policy paradigm achieves better performance. This result highlights the ability of the on-policy paradigm to identify task-relevant parameters. Building on this observation, imposing random direction constraints leads to a performance drop, whereas using the update directions identified by the on-policy paradigm achieves the best performance. This further demonstrates that the performance gains of OPSFT arise from updating parameters along directions that support generalization, rather than merely from the regularization effect of gradient masking.

\begin{table}[t]
\small
\centering
\caption{We report the performance of SFT with (i)~random update location constraints, (ii)~on-policy update location constraints, (iii)~on-policy update location constraints with random update sign constraints, and (iv)~the complete on-policy update direction constraints used in OPSFT. Experiments are conducted with Qwen3-4B on the DeepMath dataset. GRPO is selected as the on-policy paradigm that provides update location and direction constraints.}
\label{apptab:random}
\setlength{\tabcolsep}{4pt}
\begin{tabular}{@{}cc|cccc|c@{}}
\toprule
Location constraint & Direction constraint & AIME24 & AIME25 & HMMT25-Feb & HMMT25-Nov & Mean \\ \midrule
Random & - & 41.67 & 34.17 & 24.17 & 23.33 & 30.84 \\
On-policy & - & 46.67 & 40.00 & 21.67 & 27.92 & 34.07 \\
On-policy & Random & 37.50 & 31.67 & 18.33 & 23.33 & 27.71 \\
On-policy & On-policy & 52.50 & 48.33 & 31.25 & 32.50 & 41.15 \\ \bottomrule
\end{tabular}
\end{table}

\begin{figure}[t]
  \centering
  \includegraphics[width=1.0\linewidth]{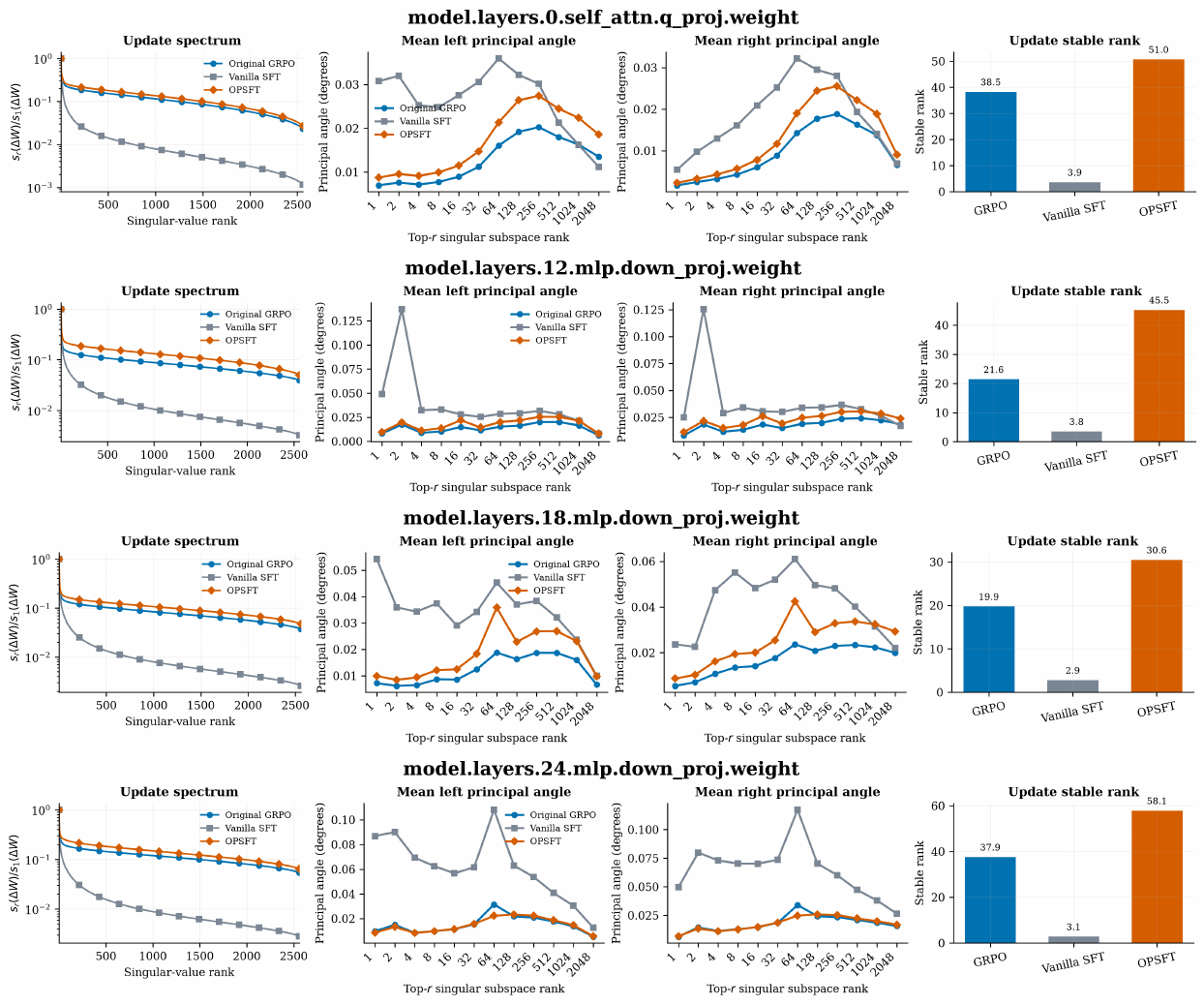}
  \caption{We provide the principal values of parameter updates, the rotation angle of parameter matrices' principal singular vectors, and the stable rank of parameter updates across different layers and methods. Experiments are conducted using Qwen3-4B trained on the DeepMath dataset.
  }
  \label{appfig:app_spectrum}
\end{figure}

\noindent\textbf{Principle components of weight matrices}. 
In Figure~\ref{appfig:app_spectrum}, we analyze the weight updates of different methods from a spectral perspective. By comparing the updates across different layers, we find that constraining SFT updates to the on-policy update direction at the weight level also makes its spectral properties more similar to those of the on-policy paradigm. First, the singular values of the update matrices in OPSFT are more evenly distributed, resembling those of GRPO, whereas the updates of vanilla SFT are concentrated in the top few singular values. This observation is further reflected by the substantially higher stable ranks of GRPO and OPSFT. Moreover, compared with SFT, which induces relatively large rotations of the principal singular vectors of the weights, OPSFT and GRPO exhibit substantially smaller rotations. The spectral similarities between OPSFT and GRPO suggest that these properties may arise as a consequence of the on-policy update direction, rather than being intrinsic to the on-policy training paradigm itself. Once the update direction is identified, even SFT can induce similar spectral behaviors.

\section{Additional Discussions}
\subsection{Discussions with Gradient Projection Methods}
OPSFT is closely related to existing projected gradient descent methods, which are typically proposed to reduce training costs~\citep{chen2019non,zhao2024galore}. From an implementation perspective, both OPSFT and these methods constrain model updates. The key difference is that we characterize the update direction through on-policy paradigms, while other methods determine the projection subspace by decomposing the SFT gradient matrix. More importantly, through gradient projection, we reveal that this direction is a key factor in transferring the strong generalization of on-policy paradigms to SFT. This finding extends the application of gradient projection beyond efficiency, demonstrating its potential for improving the generalization performance on complex reasoning tasks.

\subsection{Limitations}
\label{app:limitations}
Although the proposed OPSFT demonstrates the critical role of the on-policy update direction in the generalization performance of LLM post-training, our experiments are conducted with high-quality reasoning trajectories, \textit{i.e.}, trajectories generated by a teacher model with substantially stronger performance than the model being post-trained on the corresponding tasks. In practical scenarios, since on-policy paradigms continuously train on trajectories generated by the current policy, their trajectories may be better than those collected by SFT in terms of diversity, correctness, and their suitability for the target model~\citep{ju2025reasoning}. However, existing studies on measuring the quality and suitability of reasoning trajectories remain limited~\citep{yang2026reasoning}, preventing us from explicitly accounting for this factor in this paper. In future work, we will investigate how the properties of reasoning trajectories affect model generalization and further examine their impact on the resulting update direction and the performance of OPSFT.

\end{document}